%% file: MEIM.tex
\documentclass{article}
\usepackage{ijcai22}

\usepackage{times}
\usepackage{soul}
\usepackage{url}
\usepackage[hidelinks]{hyperref}
\usepackage[utf8]{inputenc}
\usepackage{algorithmic}
\usepackage{lipsum} 
\usepackage[disable]{todonotes} 

\usepackage{graphicx} 
\usepackage{subfigure}
\usepackage{tikz} 
\usepackage{adjustbox}  

\usepackage{booktabs} 
\usepackage{multirow} 
\usepackage{makecell} 
\usepackage{array, tabularx, boldline} 
\usepackage[small]{caption} 

\usepackage{amsmath} 
\usepackage{amssymb} 

\usepackage{amsthm}
\usepackage{amsfonts}
\usepackage{bm} 
\usepackage{mathtools} 
\usepackage{mathdots} 
\usepackage{xfrac} 
\usepackage{faktor} 
\usepackage{cancel} 

\usepackage{algorithm}

\makeatletter
\def\blfootnote{\gdef\@thefnmark{}\@footnotetext}  
\makeatother

\input{./math_commands.tex} 

\title{MEIM: Multi-partition Embedding Interaction Beyond Block Term Format for Efficient and Expressive Link Prediction}

\author{
Hung-Nghiep Tran$^1$
\and
Atsuhiro Takasu$^{1,2}$
\affiliations
$^1$National Institute of Informatics, Japan\\
$^2$The Graduate University for Advanced Studies, SOKENDAI, Japan\\
\emails
\{nghiepth, takasu\}@nii.ac.jp
}

\begin{document}

\maketitle

\begin{abstract}
	\small  
	Knowledge graph embedding aims to predict the missing relations between entities in knowledge graphs. Tensor-decomposition-based models, such as ComplEx, provide a good trade-off between efficiency and expressiveness, that is crucial because of the large size of real world knowledge graphs. The recent multi-partition embedding interaction (MEI) model subsumes these models by using the block term tensor format and provides a systematic solution for the trade-off. However, MEI has several drawbacks, some of which carried from its subsumed tensor-decomposition-based models. In this paper, we address these drawbacks and introduce the Multi-partition Embedding Interaction iMproved beyond block term format (MEIM) model, with independent core tensor for ensemble effects and soft orthogonality for max-rank mapping, in addition to multi-partition embedding. MEIM improves expressiveness while still being highly efficient, helping it to outperform strong baselines and achieve state-of-the-art results on difficult link prediction benchmarks using fairly small embedding sizes. The source code is released at \underline{\url{https://github.com/tranhungnghiep/MEIM-KGE}}. \blfootnote{\scriptsize{In the International Joint Conference on Artificial Intelligence (IJCAI), 2022.}}
\end{abstract}

\section{Introduction}
Knowledge graphs are used to represent relational information between entities. There are large real world knowledge graphs such as YAGO \cite{mahdisoltani_yago3knowledgebase_2015} containing millions of entities. These knowledge graphs and their representations can be used in artificial intelligent applications such as semantic queries and question answering \cite{tran_exploringscholarlydata_2019} \cite{tran_multirelationalembeddingknowledge_2020}. 

Knowledge graph embedding aims to predict the missing relations between entities in knowledge graphs. They usually represents a triple $ (h, t, r) $ as embeddings and use a score function to compute its matching score $ \gS(h, t, r) $. The score function defines the \textit{interaction mechanism} such as bilinear map between the embeddings and the \textit{interaction pattern} specifying how embedding entries interact with each other. 

On large real world knowledge graphs, a good tradeoff between efficiency and expressiveness is crucial. Tensor-decomposition-based models usually provide good trade-off by designing special interaction mechanisms with sparse and expressive interaction patterns. They have been subsumed by the recent multi-partition embedding interaction (MEI) model \cite{tran_multipartitionembeddinginteraction_2020}, that divides the embedding vector into multiple partitions for sparsity, automatically learns the local interaction patterns on each partition for expressiveness, then combines the local scores to get the full interaction score. The trade-off between efficiency and expressiveness can be systematically controlled by changing the partition size and learning the interaction patterns through the core tensor of the block term tensor format.

However, by looking from two perspectives beyond the scope of block term format, that is, ensemble boosting effects and max-rank relational mapping, we identify two drawbacks of MEI and related tensor-decomposition-based models. First, using one core tensor for all partitions leads to similar local interactions that may harm the ensemble effects of the full interaction in MEI. Second, the relational mapping matrices in MEI are generated from relatively small relation embeddings, that may make the model prone to degenerate. 

In this paper, we propose the novel Multi-partition Embedding Interaction iMproved beyond block term format (MEIM) model with two techniques, namely independent core tensor to improve ensemble effects and max-rank mapping by soft orthogonality to improve model expressiveness. This introduces new aspects to tensor-decomposition-based models exploiting ensemble boosting effects and max-rank relational mapping, in addition to multi-partition embedding by MEI. MEIM improves expressiveness while still being highly efficient, helping it to outperform strong baselines and achieve new state-of-the-art results on difficult link prediction benchmarks using fairly small embedding sizes. 

In general, our contributions include the following.
\begin{itemize}
	\item We propose MEIM, a novel tensor-decomposition-based model with independent core tensor and max-rank mapping by soft orthogonality to improve expressiveness.
	
	\item We extensively experiment to show that the proposed model is highly efficient and expressive, and achieves state-of-the-art results.
	
	\item We analyze the proposed model to clarify its characteristics and demonstrate the critical advantages of its soft orthogonality compared to previous models.
\end{itemize}

\begin{figure*}[ht]
	\centering
	\includegraphics[width=1.\textwidth]{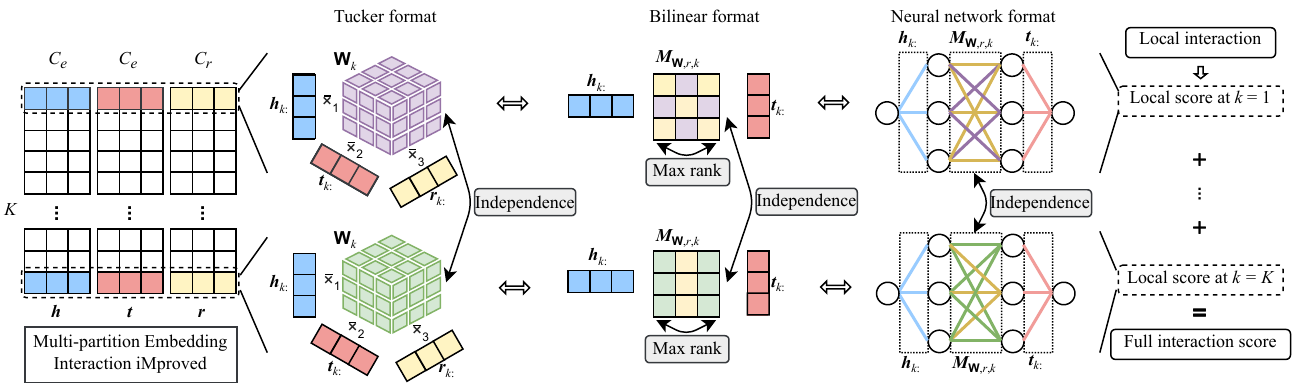}
	\caption{MEIM architecture: multi-partition embedding interaction improved with independent core tensors and max-rank mapping matrices. The new aspects are noted by the grey boxes.}
	\label{fig:meim_architecture}
\end{figure*}

\section{Background} \label{sect:background_mei}

\subsection{Notations and Definitions} \label{sect:notation} 

A knowledge graph is a collection of triples $ \gD $, with each triple denoted as a tuple $ (h, t, r) $, where $ h $ and $ t $ are head and tail entities in the entity set $ \gE $ and $ r $ belongs to the relation set $ \gR $. A knowledge graph can also be represented by a third-order binary \textit{data tensor} $ \tG \in \{0, 1\}^{|\gE| \times |\gE| \times |\gR|} $, where each entry $ g_{htr} = 1 \Leftrightarrow (h, t, r) \text{ exists in } \gD $. 

The \textit{contextual link prediction} task, also called link prediction, aims to predict the connection between two entities given a relation as the context. 

\subsection{Tensor-Decomposition-based Knowledge Graph Embedding Models}
There are several knowledge graph embedding models that adapt tensor formats to represent the knowledge graph data tensor and adapt tensor decomposition methods to solve the link prediction task. This approach has led to some of the best models in terms of efficiency and expressiveness.

RESCAL \cite{nickel_threewaymodelcollective_2011} is an early model that adapts Tucker2 decomposition to compute the embedding interaction score by a bilinear map 
\begin{align} \label{eq:rescal}
\gS(h,t,r) =\ &\vh^\top \mM_r \vt,
\end{align}
where $ \vh, \vt \in \sR^D $, and $ \mM_r \in \sR^{D \times D} $ are the embeddings of $ h $, $ t $, and $ r $, respectively, with $ D $ being the embedding size. RESCAL is expressive, but the mapping matrix $ \mM_r $ grows quadratically with embedding size, making the model expensive and prone to overfitting. 

The most simple model is DistMult \cite{yang_embeddingentitiesrelations_2015}, that uses a sparse diagonal mapping matrix $ \mM_r = \diag(\vr) $, where $ \vr \in \sR^D $ is the relation embedding vector. It is an adaptation of CP decomposition \cite{kolda_tensordecompositionsapplications_2009}, with the score function written as a trilinear product
\begin{align} \label{eq:trilinear}
\gS(h,t,r) =\ &\vh^\top \diag(\vr) \vt =\ \textstyle\sum_i h_i t_i r_i =\ \langle \vh, \vt, \vr \rangle.
\end{align}

Recent models aim to be more expressive than DistMult but still efficient by designing new special interaction mechanisms between the embeddings. It has been shown that these models are equivalent to bilinear model with sparse block-diagonal mapping matrix $ \mM_r $, 
\begin{align} \label{eq:sparsebilinear}
\mM_{r}
= 
\textstyle\begin{bmatrix}
\mM_{r,1} & 0 & 0 \\
0 & \ddots & 0 \\
0 & 0 & \mM_{r,K}
\end{bmatrix},
\end{align}
where the matrix block $ \mM_{r,k} $ have a special interaction pattern resulted from the specific interaction mechanism \cite{tran_multipartitionembeddinginteraction_2020}. For example, ComplEx \cite{trouillon_complexembeddingssimple_2016} uses complex-valued trilinear product, resulting in the 2-dimensional rotation--scaling pattern on complex plane, 
\begin{align}
\mM_{r,k} = \textstyle\begin{bmatrix} \text{real}(r)_k & - \text{imaginary}(r)_k \\ \text{imaginary}(r)_k & \text{real}(r)_k \end{bmatrix}. 
\end{align}

Sparsity and the interaction pattern are crucial concepts in previous tensor-decomposition-based models. They are generalized by the recent multi-partition embedding interaction (MEI) model \cite{tran_multipartitionembeddinginteraction_2020}, that divides the embedding vector into multiple partitions, learns the local interaction patterns on each partition, then sums the local scores to get the full interaction score. The trade-off between efficiency and expressiveness can be systematically controlled by changing the partition size and learning the interaction patterns through the core tensor of the block term tensor format.

\section{Multi-partition Embedding Interaction iMproved Beyond Block Term Format} \label{sect:model_instance} 
In this section, we propose the MEIM model that introduces new aspects to tensor-decomposition-based models exploiting \textit{ensemble boosting effects} and \textit{max-rank relational mapping}, in addition to multi-partition embedding by MEI, as illustrated in Figure \ref{fig:meim_architecture}.

\subsection{The Score Function} \label{sect:model_blockterm} 
Inspired by the generality of MEI, we adopt the multi-partition embedding representation, with Tucker format \cite{tucker_mathematicalnotesthreemode_1966} for local interaction and block term format \cite{delathauwer_decompositionshigherordertensor_2008a} for full interaction. Following convention of MEI, triple $ (h, t, r) $ provides the embedding vectors $ \vh, \vt \in \sR^{D_e} $, and $ \vr \in \sR^{D_r} $ that are treated as multi-partition embedding matrices $ \mH, \mT \in \sR^{K \times C_e} $, and $ \mR \in \sR^{K \times C_r} $, respectively. The score function of MEIM is written as
\begin{align}
\gS (h,t,r;\bm{\theta}) =\ &\textstyle\sum_{k = 1}^{K} \gS_k (h,t,r;\bm{\theta}) \label{eq:scoremeiblockterm}\\
=\ &\textstyle\sum_{k = 1}^{K} \tW_k \bar{\times}_1 \vh_{k:} \bar{\times}_2 \vt_{k:} \bar{\times}_3 \vr_{k:} \label{eq:scoremeitensorproduct}\\
=\ &\textstyle\sum_{k = 1}^{K} \vh_{k:}^\top (\tW_k \bar{\times}_3 \vr_{k:}) \vt_{k:}\\
=\ &\textstyle\sum_{k = 1}^{K} \vh_{k:}^\top \mM_{\tW, r, k} \vt_{k:}, \label{eq:scoremeibilinear}
\end{align}
where $ \vh_{k:} $, $ \vt_{k:} $, and $ \vr_{k:} $ are the embedding partitions $ k $ \footnote{Partitions are column vectors, transpose notation is omitted for simplicity. Illustration as row is just for easy visualization.}; $ \tW_k \in \sR^{C_e \times C_e \times C_r} $ is the core tensor at partition $ k $; $ \bar{\times}_n $ denotes the $ n $-mode tensor product with a vector; and $ \mM_{\tW, r, k} \in \sR^{C_e \times C_e} $ is the bilinear mapping matrix. 

Note that Eq. \ref{eq:scoremeiblockterm} shows the sum of local interaction scores $ \gS_k (h,t,r;\bm{\theta}) $. Eq. \ref{eq:scoremeitensorproduct} shows the block term format. Eq. \ref{eq:scoremeibilinear} shows the bilinear format with block-diagonal mapping matrix, in which each block $ \mM_{\tW, r, k} $ is generated by $ \tW_k \bar{\times}_3 \vr_{k:} $. Eq. \ref{eq:scoremeibilinear} can be seen as a dynamic linear neural network \cite{tran_multipartitionembeddinginteraction_2020}, where the hidden layer $ \mM_{\tW, r, k} $ is generated by another hyper neural network \cite{ha_hypernetworks_2016} with the core tensor $ \tW_k $ as its weights and $ \vr_{k:} $ as its input.

The main novelties of MEIM are in better parameterization of the core tensor and mapping matrix, by looking from two perspectives beyond the scope of block term format, that is, ensemble boosting effects and max-rank relational mapping.

\subsection{Core Tensor for Ensemble Boosting} \label{sect:model_nonsharedcore} 
Technically, the score function in Eq. \ref{eq:scoremeiblockterm} can be seen as an ensemble system of $ K $ local interactions by summing their scores. This works in a similar manner to gradient boosting \cite{mason_boostingalgorithmsgradient_1999} because the full interaction score is computed at training time and gradients are back-propagated to optimize all local interactions together, enabling them to implicitly minimize the residual error of each other. 


The MEI model briefly mentioned this perspective but did not actively exploit the ensemble boosting effects. MEI used only one core tensor for all partitions by enforcing the following \textit{shared core tensor} constraint:
\begin{align}
\tW_{1} = \tW_{2} = \dots = \tW_{K} = \tW, \label{eq:constraintsharedcore}
\end{align}
where $ \tW \in \sR^{C \times C \times C} $ is the learned shared core tensor, although block term format may have different core tensors. MEI learns one interaction pattern for all partitions, similar to that of other tensor-decomposition-based models such as ComplEx. However, this may be harmful to the performance of an ensemble system because of the similar local interaction patterns instead of combining different patterns. To exploit the ensemble boosting effects, it is crucial to promote independence and difference between the local interactions. 

\subsubsection{Independent Core Tensor Parameterization}
There are several potential ways to resolving the problem of independence and difference between the local interactions. First, we may enforce the predicted scores to be different:
\begin{align}
\forall k, l \in \{1, \dots, K\}, k \neq l: \gS_k (\cdot, \cdot, \cdot) \neq \gS_l (\cdot, \cdot, \cdot). \label{eq:constraintnonsharedscore}
\end{align}
However, because this forces the predicted scores to be not the same, the local interactions cannot be correct at the same time. Thus, this is an overstrong constraint that may cause more harm than good. Another way is to explicitly enforce different parameters of the local interaction functions, such as the core tensors:
\begin{align}
\forall k, l \in \{1, \dots, K\}, k \neq l: \tW_{k} &\neq \tW_{l}. \label{eq:constraintnonsharedcore}
\end{align}
However, such explicit difference constraints do not guarantee different interactions because of symmetric swapping or scaling reparameterizations of the core tensors and the embeddings. For example, scaling the core tensor by $ x $ and the corresponding embedding by $ 1/x $ do not change the score. 

Therefore, in this paper, we do not use difference constraints, but instead independence constraints, denoted as:
\begin{align}
	\tW_{1} \perp \tW_{2} \perp \dots \perp \tW_{K}. \label{eq:constraintindependentcore}
\end{align}
For simplicity, we let the model implicitly learn \textit{independent core tensors} from data automatically by removing the constraint in Eq. \ref{eq:constraintsharedcore}. Using independent core tensors enables the model to learn independent and possibly different local interactions to improve the ensemble boosting effects. More advanced methods based on further studying the ensemble boosting effects are left for future work. 

Note that MEI is easy to implement in common deep learning frameworks by using only matrix product. However, independent core tensor makes the multi-partition tensor product in MEIM more difficult to implement efficiently. To resolve this problem, we stack together $ K $ independent core tensors as a fourth-order tensor:
\begin{align}
\tW = &\left[\tW_{1}, \tW_{2}, \dots, \tW_K\right], \label{eq:nonsharedcore}
\end{align}
where $ \tW \in \sR^{K \times C \times C \times C} $ is used for the Einstein sum notation technique to create an efficient implementation. 

\subsection{Max-Rank Relational Mapping} \label{sect:model_mapping} 



MEI is a \textit{contextual link prediction} model with two components, the multi-partition relation-based contextual mapping and the simple dot product matching. The MEI model used a linear mapping where $ \mM_{\tW, r, k} $ can be any $ C_e \times C_e $ square matrix. However, when some columns of $ \mM_{\tW, r, k} $ are dependent, it becomes a singular matrix with rank smaller than $ C_e $. Such singular matrix would map the head embedding to a subspace with lower dimension, thus reduce the embedding space effective size. It may make the score function degenerate to always produces score $ 0 $ for some entities, thus, drastically reduces the expressiveness of the model. This may become a critical issue in MEI because $ \mM_{\tW, r, k} $ is generated from a relatively small relation embedding partition $ \vr_{k:} $ as shown in Eq. \ref{eq:scoremeibilinear}. To resolve this problem, the mapping matrices $ \mM_{\tW, r, k} $ need to have max rank, that may not always be full rank $ C_e $, but should be as large as possible. 

\subsubsection{Max-Rank Mapping Matrix by Soft Orthogonality}
There are several different types of full rank matrices. A particularly interesting case is the orthogonal matrices, that act as the rotational transformations when the determinant is $ 1 $ and an additional reflection across the origin when the determinant is $ -1 $. These matrices have linearly independent column vectors with unit Frobenius norms, written as the constraints:
\begin{align}
	\forall k \in \{1, \dots, K\}: \mM_{\tW, r, k}^\top \mM_{\tW, r, k} = \mI, \label{eq:constraintorthohard}
\end{align}
where $ \mI \in \sR^{C_e \times C_e} $ is the identity matrix. In addition, note that the mapping matrices are generated by the core tensors and the relation embedding vectors. To reduce the search space and make optimization easier, we also constrain each relation partition $ \vr_{k:} $ to have unit norm:
\begin{align}
	\forall k \in \{1, \dots, K\}: \vr_{k:}^\top \vr_{k:} = 1. \label{eq:constraintunitnormhard}
\end{align}

Because of the way $ \mM_{\tW, r, k} $ are generated, it may be difficult to enforce the mapping matrices to be strictly orthogonal. Even if it is possible, it may not always be preferable because the main objective is link prediction. Geometrically speaking, the orthogonal matrices form the Stiefel manifold \cite{sokol_informationgeometryorthogonal_2020}, we need to make sure that the mapping matrices stay as close to this manifold as possible without hurting the link prediction accuracy. Using Lagrangian relaxation, the hard constraints in Eq. \ref{eq:constraintorthohard} and Eq. \ref{eq:constraintunitnormhard} can be converted to the following soft orthogonality loss term:
\begin{align}
\gL_{\text{ortho}} = \lambda_{\text{ortho}} &\bigl( \textstyle\sum_{k = 1}^{K} || \mM_{\tW, r, k}^\top \mM_{\tW, r, k} - \mI ||_2^2. \nonumber\\
&\ \ + \lambda_{\text{unitnorm}} \textstyle\sum_{k = 1}^{K} |\vr_{k:}^\top \vr_{k:} - 1|^p \bigr), \label{eq:constraintorthogonalityunitnorm}
\end{align}
where $ \lambda_{\text{ortho}} $, $ \lambda_{\text{unitnorm}} $, and $ p $ are hyperparameters. To penalize large differences from unit norm, $ p=3 $ is used. When $ \lambda_{\text{ortho}} $ is small, the model tends to learn linear mappings. When it is large, the model tends to learn strictly orthogonal mappings. 

The soft orthogonality loss enables the model to balance between orthogonality and link prediction. Minimizing this loss effectively influences the column vectors to be orthogonal and have unit norms, pushes the matrix $ \mM_{\tW, r, k} $ close to the Stiefel manifold and maximizes its rank in balance with the link prediction objective. 

\subsection{Connections to Previous Models} \label{sect:connection_advantage}
Compared to the MEI model, MEIM inherits its key benefit, that is, the systematic trade-off between efficiency and expressiveness by controlling the partition size and learning the interaction patterns. About computational cost, the number of parameters in MEIM is similar to that of MEI, except that the independent core tensors use $ K C_e^2 C_r $ instead of $ C_e^2 C_r $ parameters, but this becomes less relevant when the number of entities and relations are large, and MEIM is as fast as MEI by using Einstein sum notation technique for implementation of independent core tensor product. 

On the other hand, MEIM introduces new parameterization techniques to address some major drawbacks of MEI. With independent core tensor, MEIM can learn independent local interaction patterns to improve ensemble effects. With soft orthogonality, MEIM balances the orthogonality and link prediction objectives to learn max-rank mapping that improves expressiveness of the model.

Note that, coincidentally, some knowledge graph embedding models also try to use orthogonal mapping, not to maximize the mapping matrix rank but to model some relational patterns, such as symmetry, inversion, and composition \cite{sun_rotateknowledgegraph_2019}. There are many different ways to design and use orthogonal mapping in knowledge graph embedding. For example, RotatE \cite{sun_rotateknowledgegraph_2019} uses rotation in the complex plane, QuatE \cite{zhang_quaternionknowledgegraph_2019} uses rotation in the quaternion space, GC-OTE \cite{tang_orthogonalrelationtransforms_2020} uses rotation by the Gram Schmidt process, and RotH \cite{chami_lowdimensionalhyperbolicknowledge_2020} uses rotation by the Givens matrix. However, there are some important differences to our method. First, all of them use rigid orthogonality where the mapping matrix is strictly orthogonal. This may be suboptimal because orthogonal mapping may not always be useful for link prediction, and even when it is useful, strictly orthogonal mapping may reduce the potential result. In contrast, we approach the problem from the perspective of maximizing the mapping matrix rank, thus, we use soft orthogonality as a means instead of rigid orthogonality. Soft orthogonality can avoid the above problems because we can choose how to balance the orthogonality and the link prediction objectives for each dataset, as we will show this is crucial for knowledge graph embedding in experiments. Second, MEIM has a totally different model architecture compared to previous orthogonal models, because the mapping matrix in MEIM is not learned directly but generated. Finally, MEIM is a semantic matching model, with the matching score computed by inner product, whereas other orthogonal models are translation-based model, in which the score is computed by distance.

\section{Loss Function and Learning} \label{sect:learning} 
The model is learned via the link prediction task that can be modeled as a multi-class classification problem. There are two directions, classifying correct tail entity $ t $ among all entities given a $ (h, r) $ pair, and classifying correct head entity $ h $ among all entities given a $ (t, r) $ pair. Following recent practice, we use the \textit{1-vs-all} and \textit{k-vs-all} sampling methods with the softmax cross-entropy loss function \cite{dettmers_convolutional2dknowledge_2018} \cite{ruffinelli_youcanteach_2020}. 

First, using \textit{1-vs-all} and \textit{k-vs-all} sampling, we define the groundtruth categorical distributions over all entities $ \gE $ given $ (h, r) $ and given $ (t, r) $, denoted $ \hat{p}_{h\dot{t}r} $ and $ \hat{p}_{\dot{h}tr} $, respectively. Second, we compute the corresponding predicted distributions $ p_{h\dot{t}r} $ and $ p_{\dot{h}tr} $ using the softmax function on the matching scores. The link prediction loss is the cross-entropy summed over training data $ \gD $:
\begin{align} \label{eq:loss_fullcrossentropy}
\gL_{\text{link\_prediction}} =\ - \textstyle\sum_{(h, t, r) \in \gD} \bigl( &\textstyle\sum_{\dot{t} \in \gE} \hat{p}_{h\dot{t}r} \log p_{h\dot{t}r} \nonumber\\
+ &\textstyle\sum_{\dot{h} \in \gE} \hat{p}_{\dot{h}tr} \log p_{\dot{h}tr} \bigr).
\end{align} 

The final loss function is the sum of link prediction loss and soft orthogonality loss: 
\begin{align} \label{eq:loss}
\gL = \gL_{\text{link\_prediction}} + \gL_{\text{ortho}},
\end{align} 
with $ \lambda_{\text{ortho}} $ tuned to balance the two loss terms. For regularization, we only used dropout \cite{srivastava_dropoutsimpleway_2014} and batch normalization \cite{ioffe_batchnormalizationaccelerating_2015}. All parameters in the model are learned in an end-to-end fashion by minimizing the loss using mini-batch stochastic gradient descent with Adam optimizer \cite{kingma_adammethodstochastic_2015}. 

We did not use some recent complemental techniques for training, regularization, and additional losses that can be used with existing methods including MEIM. Adding them to MEIM to further improve the result is left for future work.

\section{Experiments} \label{sect:experiment}
\subsection{Experimental Settings} \label{sect:expsetting} 
\paragraph{Datasets.} We use three standard benchmark datasets, as shown in Table \ref{tab:data}. WN18RR \cite{dettmers_convolutional2dknowledge_2018} is a subset of WordNet containing lexical information. FB15K-237 \cite{toutanova_observedlatentfeatures_2015} is a subset of Freebase containing general facts. In addition, YAGO3-10 \cite{mahdisoltani_yago3knowledgebase_2015} is a large and very competitive dataset containing general facts from Wikipedia.

\begin{table}
	\centering
	\begin{adjustbox}{max width=\columnwidth}
		\begin{tabular}{@{\extracolsep{-8pt}}lcccccc}
			Dataset & $ |\gE| $ & $ |\gR| $ & Train & Valid & Test & Avg. degree\\ 
			\midrule
			
			WN18RR & 40,943 & 11 & 86,835 & 3,034 & 3,134 & 2.12 \\
			FB15K-237 & 14,541 & 237 & 272,115 & 17,535 & 20,466 & 18.71 \\
			YAGO3-10 & 123,182 & 37 & 1,079,040 & 5,000 & 5,000 & 8.76 \\
			
			\bottomrule
		\end{tabular}
	\end{adjustbox}
	\caption{Datasets statistics.}
	\label{tab:data}
\end{table}

\paragraph{Evaluations.} We evaluate on the link prediction task. For each true triple $ (h, t, r) $ in the test set, we replace $ h $ and $ t $ by every other entity to generate corrupted triples $ (h', t, r) $ and $ (h, t', r) $, respectively. The model tries to rank the true triple $ (h, t, r) $ before the corrupted triples based on the score $ \gS $. We compute $ MRR $ (mean reciprocal rank) and $ H@k $ for $ k \in \{1, 3, 10\} $ (how many triples correctly ranked in the top $ k $) \cite{trouillon_complexembeddingssimple_2016}. The higher $ MRR $ and $ H@k $ are, the better the model performs. Filtered metrics are used to avoid penalization when ranking other true triples before the current target triple \cite{bordes_translatingembeddingsmodeling_2013}.

Note that MEIM can solve other tasks by converting them to \textit{contextual link prediction} task. For example, by defining the \textit{alignment relation} $ r_a $, we have the alignment triples of the form $ (e_1, e_2, r_a) $, that can be used directly in MEIM for solving entity alignment. These tasks are left for future work.

\paragraph{Implementations.} The MEIM model is implemented as a neural network using PyTorch. Following MEI, we use embedding size $ K=3, C=100 $ on WN18RR and FB15K-237, and $ K=5, C=100 $ on YAGO3-10. They are equivalent in terms of model size to ComplEx with embedding size $ 190 $ on WN18RR, $ 250 $ on FB15K-237, and $ 270 $ on YAGO3-10, that are relatively small compared to most related work. By preliminary experiments, we use \textit{k-vs-all} sampling on WN18RR and \textit{1-vs-all} sampling on other datasets. We use batch size $ 1024 $ and learning rate 3e-3, with exponential decay $ 0.995 $ on YAGO3-10 and $ 0.99775 $ on other datasets. Dropout and batch normalization are used on the input and hidden layers, that is, on $ \vh $ and $ \vh^\top \mM_r $, respectively. Other hyperparameters are tuned by grid search, with grids $ [0, 0.75] $ and step size $ 0.01 $ for drop rates, \{1, 1e-1, 1e-2, 1e-3, 1e-4, 1e-5, 0\} for $ \lambda_{\text{ortho}} $, \{1, 1e-1, 1e-2, 1e-3, 5e-4, 1e-4, 0\} for $ \lambda_{\text{unitnorm}} $. The following values were found. On WN18RR, input drop rate is $ 0.71 $, hidden drop rate is $ 0.67 $, $ \lambda_{\text{ortho}} = $ 1e-1, $ \lambda_{\text{unitnorm}} = $ 5e-4. On FB15K-237, input drop rate is $ 0.66 $, hidden drop rate is $ 0.67 $, $ \lambda_{\text{ortho}} = $ 0, $ \lambda_{\text{unitnorm}} = $ 0. On YAGO3-10, input drop rate is $ 0.1 $, hidden drop rate is $ 0.15 $, $ \lambda_{\text{ortho}} = $ 1e-3, $ \lambda_{\text{unitnorm}} = $ 0. Hyperparameters were tuned to maximize the validation MRR.

\paragraph{Baselines.} The main baselines are MEI and tensor-decomposition-based models such as ComplEx, RotatE, QuatE that we aim to improve. MEI results are reproduced in this paper using comparable model sizes and recent training techniques that lead to better performance than previously reported. We also compare to translation-based models such as TransE, neural-network-based models such as ConvE, and recent models such as RotH. We tried to evaluate against popular and strong baselines with comparable settings and evaluation protocol for fair and informative comparisons. These exclude incomparable results that use extra data, excessively large model size, complemental techniques, or inappropriate evaluation protocol \cite{sun_reevaluationknowledgegraph_2020}. Note that even at normal model size, MEIM can get state-of-the-art results.

\subsection{Parameter Efficiency}
We first compare the results and model sizes of MEIM to that of popular baselines as shown in Table \ref{tab:param_efficiency}. We can see that MEIM strongly outperforms these baselines while using roughly similar number of parameters. 

\begin{table}[t]
	\centering  
	\begin{adjustbox}{max width=\linewidth}
		\begin{tabular}{@{\extracolsep{-8pt}}crclrclcc}
			
			& \multicolumn{2}{c}{WN18RR} && \multicolumn{2}{c}{FB15K-237} && \multicolumn{2}{c}{YAGO3-10} \\
			\cmidrule{2-3} \cmidrule{5-6} \cmidrule{8-9}
			& \#params & MRR && \#params & MRR && \#params & MRR \\
			\cmidrule{1-3} \cmidrule{5-6} \cmidrule{8-9}
			
			TuckER & 9.4 M & 0.470 && 11.0 M & 0.358 && -- & -- \\
			RotatE & 41.0 M & 0.476 && 14.8 M & 0.338 && 61.6 M & 0.495 \\
			ComplEx-N3 & 20.5 M & 0.480 && 7.4 M & 0.357 && 61.6 M & 0.569 \\
			RotH & 20.5 M & 0.496 && 7.4 M & 0.344 && 61.6 M & 0.570 \\
			
			
			MEI & 15.7 M & 0.481 && 7.4 M & 0.365 && 66.5 M & 0.578 \\
			
			\midrule
			
			MEIM & 15.3 M & \textbf{0.499} && 7.4 M & \textbf{0.369} && 66.6 M & \textbf{0.585} \\
			
		\end{tabular}
	\end{adjustbox}
	\caption[]{Parameter efficiency compared to popular baselines.}
	\label{tab:param_efficiency}
\end{table}

In addition, Table \ref{tab:param_scale} shows the scalability of MEIM and other baselines at different model sizes. MEIM outperforms the baselines at every model size and achieves the best result overall. This demonstrates that MEIM achieves a better balance between efficiency and expressiveness.

\begin{table}[t]
	\centering  
	\begin{adjustbox}{max width=\linewidth}
		\begin{tabular}{@{\extracolsep{-4pt}}rcclcclcc}
			
			& \multicolumn{2}{c}{ComplEx-N3} && \multicolumn{2}{c}{MEI} && \multicolumn{2}{c}{MEIM} \\
			\cmidrule{2-3} \cmidrule{5-6} \cmidrule{8-9}
			\#params & MRR & H@10 && MRR & H@10 && MRR & H@10 \\
			\cmidrule{1-3} \cmidrule{5-6} \cmidrule{8-9}
			
			1.2 M & 0.404 & 0.439 && 0.448 & 0.500 && 0.460 & 0.525 \\ 
			3.8 M & 0.453 & 0.507 && 0.476 & 0.544 && 0.488 & 0.557 \\ 
			6.5 M & 0.464 & 0.528 && 0.481 & 0.547 && 0.494 & 0.569 \\ 
			15.3 M & 0.473 & 0.550 && 0.481 & 0.544 && 0.499 & 0.574 \\ 
			
		\end{tabular}
	\end{adjustbox}
	\caption[]{Model scalability on WN18RR validation set.}
	\label{tab:param_scale}
\end{table}

\begin{table*}[ht]
	
	\centering  
	\begin{adjustbox}{max width=\textwidth}
		\begin{tabular}{@{\extracolsep{-4pt}}lcccclcccclcccc}
			
			& \multicolumn{4}{c}{\textbf{WN18RR}} && \multicolumn{4}{c}{\textbf{FB15K-237}} && \multicolumn{4}{c}{\textbf{YAGO3-10}} \\
			\cmidrule(lr){2-5} \cmidrule(lr){7-10} \cmidrule(lr){12-15}
			& MRR & H@1 & H@3 & H@10 && MRR & H@1 & H@3 & H@10 && MRR & H@1 & H@3 & H@10 \\
			
			\cmidrule(lr){1-5} \cmidrule(lr){7-10} \cmidrule(lr){12-15}
			
			TransE \cite{bordes_translatingembeddingsmodeling_2013} $ ^\ddagger $ & 0.222 & 0.031 & -- & 0.524 && 0.310 & 0.217 & -- & 0.497 && 0.501 & 0.406 & -- & 0.674 \\
			TorusE \cite{ebisu_generalizedtranslationbasedembedding_2019} $ ^\sharp $ & 0.452 & 0.422 & 0.464 & 0.512 && 0.305 & 0.217 & 0.335 & 0.484 && 0.342 & 0.274 & -- & 0.474 \\
			\cmidrule(lr){1-5} \cmidrule(lr){7-10} \cmidrule(lr){12-15}
			
			ConvE \cite{dettmers_convolutional2dknowledge_2018} $ ^\sharp $ & 0.43 & 0.40 & 0.44 & 0.52 && 0.325 & 0.237 & 0.356 & 0.501 && 0.488 & 0.399 & -- & 0.658 \\
			InteractE \cite{vashishth_interacteimprovingconvolutionbased_2020} & 0.463 & 0.430 & -- & 0.528 && 0.354 & 0.263 & -- & 0.535 && 0.541 & 0.462 & -- & 0.687 \\
			CompGCN \cite{vashishth_compositionbasedmultirelationalgraph_2020} & 0.479 & 0.443 & 0.494 & 0.546 && 0.355 & 0.264 & 0.390 & 0.535 && -- & -- & -- & -- \\
			RAGAT \cite{liu_ragatrelationaware_2021} & 0.489 & \underline{0.452} & 0.503 & 0.562 && \underline{0.365} & \underline{0.273} & 0.401 & 0.547 && -- & -- & -- & -- \\
			\cmidrule(lr){1-5} \cmidrule(lr){7-10} \cmidrule(lr){12-15}
			
			RESCAL \cite{nickel_threewaymodelcollective_2011} $ ^\dagger $ & 0.467 & -- & -- & 0.517 && 0.357 & -- & -- & 0.541 && -- & -- & -- & -- \\
			DistMult \cite{yang_embeddingentitiesrelations_2015} $ ^\ddagger $ & 0.465 & 0.432 & -- & 0.532 && 0.313 & 0.224 & -- & 0.490 && 0.501 & 0.413 & -- & 0.661 \\
			ComplEx-N3 \cite{trouillon_complexembeddingssimple_2016} $ ^\nmid $ & 0.480 & 0.435 & 0.495 & 0.572 && 0.357 & 0.264 & 0.392 & 0.547 && 0.569 & 0.498 & 0.609 & 0.701 \\
			RotatE \cite{sun_rotateknowledgegraph_2019} & 0.476 & 0.428 & 0.492 & 0.571 && 0.338 & 0.241 & 0.375 & 0.533 && 0.495 & 0.402 & 0.550 & 0.670 \\
			QuatE \cite{zhang_quaternionknowledgegraph_2019} & 0.488 & 0.438 & 0.508 & 0.582 && 0.348 & 0.248 & 0.382 & 0.550 && -- & -- & -- & -- \\
			
			GC-OTE \cite{tang_orthogonalrelationtransforms_2020} & 0.491 & 0.442 & 0.511 & \underline{0.583} && 0.361 & 0.267 & 0.396 & 0.550 && -- & -- & -- & -- \\
			RotH \cite{chami_lowdimensionalhyperbolicknowledge_2020} & \underline{0.496} & 0.449 & \underline{0.514} & \textbf{0.586} && 0.344 & 0.246 & 0.380 & 0.535 && 0.570 & 0.495 & 0.612 & 0.706 \\
			
			TuckER \cite{balazevic_tuckertensorfactorization_2019} & 0.470 & 0.443 & 0.482 & 0.526 && 0.358 & 0.266 & 0.394 & 0.544 && -- & -- & -- & -- \\ 
			
			
			
			MEI \cite{tran_multipartitionembeddinginteraction_2020} $ ^\star $ & 0.481 & 0.444 & 0.496 & 0.551 && \underline{0.365} & 0.271 & \underline{0.402} & \underline{0.552} && \underline{0.578} & \underline{0.505} & \underline{0.622} & \underline{0.709} \\ 
			
			\midrule

			MEIM & \textbf{0.499} & \textbf{0.458} & \textbf{0.518} & 0.577 && \textbf{0.369} & \textbf{0.274} & \textbf{0.406} & \textbf{0.557} && \textbf{0.585} & \textbf{0.514} & \textbf{0.625} & \textbf{0.716} \\ 
			
			
		\end{tabular}
	\end{adjustbox}
	\caption[Main link prediction results.]{Link prediction results on WN18RR, FB15K-237, and YAGO3-10. $ ^\dagger $ are reported in \cite{ruffinelli_youcanteach_2020}, $ ^\nmid $ in \cite{chami_lowdimensionalhyperbolicknowledge_2020}, $ ^\ddagger $ in \cite{rossi_knowledgegraphembedding_2021}, $ ^\sharp $ YAGO3-10 in \cite{rossi_knowledgegraphembedding_2021}, $ ^\star $ are reproduced here, other results are reported in their papers.}
	\label{tab:result_hard}
\end{table*}

\subsection{Main Results} \label{sect:result}
The main results for link prediction are shown in Table \ref{tab:result_hard}. In general, MEIM achieves good result on all three datasets, including the very large and difficult YAGO3-10 dataset. Considering the most important and robust metric MRR, MEIM outperforms all baselines by a large margin on all datasets. 

The strongest baselines on FB15K-237 and YAGO3-10 include the MEI model. This demonstrates that MEI was a strong model, but MEIM successfully addresses the drawbacks of MEI and its subsumed tensor-decomposition-based models to significantly improve the result on all datasets. 

Another strong baseline is the RotH model on WN18RR, but the results on FB15K-237 are quite weak. We notice that most baselines usually have good results on one dataset and poor results on other datasets. This sets MEIM apart as it can achieve good results on all three datasets. 

\subsection{Analyses and Discussions}
Now we look into the characteristics of the model to understand why it gives good result.

\subsubsection{Ablation Study}
Table \ref{tab:ablation_model} shows the main ablation results of MEIM, in which each feature is removed from the full model to see how much it contributes to the final results. For comparison, we also ablate the \textit{multi-partition embedding} inherited from MEI. 

\begin{table}[t]
	\centering  
	\begin{adjustbox}{max width=\linewidth}
		\begin{tabular}{@{\extracolsep{-8pt}}llcc}
			
			& Ablation & MRR & H@10 \\
			\cmidrule{2-4}
			
			\multirow{5}{*}{WN18RR} & Full model & \textbf{0.499} & \textbf{0.574} \\ 
			\cmidrule{3-4}
			& without \textit{Multi-partition Embedding} & 0.492 & 0.572 \\ 
			& without \textit{Independent Core Tensor} & 0.498 & 0.573 \\ 
			& without \textit{Soft Orthogonality} & 0.483 & 0.550 \\ 
			
			\midrule
			
			\multirow{4}{*}{FB15K-237} & Full model & \textbf{0.375} & \textbf{0.562} \\ 
			\cmidrule{3-4}
			& without \textit{Multi-partition Embedding} & 0.372 & 0.554 \\ 
			& without \textit{Independent Core Tensor} & 0.370 & 0.554 \\ 
			
			\midrule
			
			\multirow{5}{*}{YAGO3-10} & Full model & \textbf{0.582} & \textbf{0.710} \\ 
			\cmidrule{3-4}
			& without \textit{Multi-partition Embedding} & 0.580 & 0.706 \\ 
			& without \textit{Independent Core Tensor} & 0.581 & 0.708 \\ 
			& without \textit{Soft Orthogonality} & 0.574 & 0.708 \\ 
			
		\end{tabular}
	\end{adjustbox}
	\caption[]{Model ablation result on the validation sets.}
	\label{tab:ablation_model}
\end{table}

In general, we can see that removing the features decreases the results, which confirms their effectiveness. In particular, the \textit{independent core tensor} works well on all three datasets. It strongly contributes to the full model results on FB15K-237, and slightly but consistently contributes to the final results on WN18RR and YAGO3-10. The effects of \textit{soft orthogonality} is more complicated. On WN18RR and YAGO3-10, soft orthogonality is crucial to achieve the best results. However, it is harmful on FB15K-237 and was not included in the full model. This mixed effect guarantees further detailed investigation as follows.

\subsubsection{The Effects of Soft Orthogonality}
Table \ref{tab:ablation_ortho} shows the results when $ \lambda_{\text{ortho}} $ is changing from no orthogonality to strong orthogonality. We see that strong orthogonality works well on WN18RR, average orthogonality works well on YAGO3-10, but none orthogonality works on FB15K-237. These results empirically demonstrate the weakness of previous models compared to MEIM. 

\begin{table}[t]
	\centering  
	\begin{adjustbox}{max width=\linewidth}
		\begin{tabular}{@{\extracolsep{-2pt}}ccclcclcc}
			
			& \multicolumn{2}{c}{WN18RR} && \multicolumn{2}{c}{FB15K-237} && \multicolumn{2}{c}{YAGO3-10} \\
			\cmidrule{2-3} \cmidrule{5-6} \cmidrule{8-9}
			$ \lambda_{\text{ortho}} $ & MRR & H@10 && MRR & H@10 && MRR & H@10 \\
			\cmidrule{1-3} \cmidrule{5-6} \cmidrule{8-9}
			
			0 & 0.483 & 0.550 && \textbf{0.375} & \textbf{0.562} && 0.574 & 0.708 \\ 
			1e-5 & 0.482 & 0.548 && 0.374 & 0.561 && 0.581 & \textbf{0.711} \\ 
			1e-4 & 0.483 & 0.553 && 0.371 & 0.556 && 0.581 & 0.709 \\ 
			1e-3 & 0.484 & 0.554 && 0.372 & 0.559 && \textbf{0.582} & 0.710 \\ 
			1e-2 & 0.497 & \textbf{0.577} && 0.362 & 0.546 && \textbf{0.582} & 0.710 \\ 
			1e-1 & \textbf{0.499} & 0.574 && 0.360 & 0.541 && 0.581 & 0.706 \\ 
			1e0 & 0.497 & 0.568 && 0.355 & 0.532 && 0.576 & 0.695 \\ 
			
		\end{tabular}
	\end{adjustbox}
	\caption[]{The effects of soft orthogonality on the validation sets.}
	\label{tab:ablation_ortho}
\end{table}

For example, models with rigid orthogonality such as QuatE and RotH get strong results on WN18RR but weak on FB15K-237, whereas models without orthogonality such as RESCAL, TuckER, and MEI get strong results on FB15K-237 but weak on WN18RR. In contrast, MEIM with soft orthogonality can get strong results on both WN18RR and FB15K-237. More importantly, the large and difficult YAGO3-10 dataset requires average orthogonality. Thus, only MEIM can get the best results on this dataset. This shows a crucial advantage of MEIM, because its soft orthogonality can be tuned to work optimally on each dataset.

\section{Related Work} \label{sect:relatedwork} 
Knowledge graph embedding methods can be categorized based on how they compute the matching score. 

\textit{Tensor-decomposition-based models} adapt tensor representation formats such as CP format, Tucker format, and block term format to represent the knowledge graph data tensor \cite{kolda_tensordecompositionsapplications_2009}. They make up most of recent state-of-the-art models such as ComplEx \cite{trouillon_complexembeddingssimple_2016}, SimplE \cite{kazemi_simpleembeddinglink_2018}, TuckER \cite{balazevic_tuckertensorfactorization_2019}, and MEI \cite{tran_multipartitionembeddinginteraction_2020}. They often achieve good trade-off between efficiency and expressiveness. Especially, the recent MEI model provides a systematic solution for this trade-off. However, they have some limitations affecting the ensemble effects and making them prone to degenerate, that we address in this paper.

\textit{Neural-network-based models} use a neural network to compute the matching score, such as ConvE \cite{dettmers_convolutional2dknowledge_2018} using convolutional neural networks, CompGCN \cite{vashishth_compositionbasedmultirelationalgraph_2020} using graph convolutional networks. These models are generally more expensive but not always get better results than tensor-decomposition-based models.

\textit{Translation-based models} use geometrical distance to compute the score, with relation embeddings act as the translation vectors, such as TransE \cite{bordes_translatingembeddingsmodeling_2013}. These models are efficient and intuitive, but they have limitations in expressive power \cite{kazemi_simpleembeddinglink_2018}. 

There are several ways to use orthogonality in knowledge graph embedding, such as RotatE \cite{sun_rotateknowledgegraph_2019} using complex product, Quaternion \cite{tran_analyzingknowledgegraph_2019} and QuatE \cite{zhang_quaternionknowledgegraph_2019} using quaternion product, GC-OTE \cite{tang_orthogonalrelationtransforms_2020} using the Gram Schmidt process, and RotH \cite{chami_lowdimensionalhyperbolicknowledge_2020} using the Givens matrix. These models usually compute the score using distance, differently from MEIM. Moreover, they use rigid orthogonality that is harmful on some datasets and potentially suboptimal on other datasets compared to soft orthogonality.

\section{Conclusion} \label{sect:conclusion} 
In this paper, we introduce new aspects to tensor-decomposition-based models exploiting ensemble boosting effects and max-rank relational mapping, in addition to multi-partition embedding by MEI. We propose the MEIM model with two new techniques, namely independent core tensor to improve ensemble effects and max-rank mapping by soft orthogonality to improve expressiveness. MEIM achieves state-of-the-art link prediction results, including the large and difficult YAGO3-10 dataset, using fairly small embedding sizes. Moreover, we analyze and demonstrate the limitations of previous rigid orthogonality models and show that MEIM with soft orthogonality works well on multiple datasets. 

For future work, it is promising to continue research on combining deep learning techniques and tensor decomposition. It is also interesting to use MEIM to solve tasks such as entity alignment and recommendation by converting them to contextual link prediction.

\section*{Acknowledgments}
This work was supported by the Cross-ministerial Strategic Innovation Promotion Program (SIP) Second Phase, ``Big-data and AI-enabled Cyberspace Technologies'' by the New Energy and Industrial Technology Development Organization (NEDO).

\bibliographystyle{named}

\input{MEIM.bbl}
\clearpage
\appendix

\section{Extra Experiments}
In this section, we present some extra experimental results that may be useful but did not fit in the conference page limit.

\subsection{Detailed Link Prediction Results on WN18RR}
In addition to the overall result, we look into the detailed relation result on WN18RR in Table \ref{tab:result_detailrelation}. In general, MEI and MEIM outperform RotatE on all relations in WN18RR. MEIM achieves best results for almost all relations, and comes closely second on a few others. This shows that MEIM works consistently well on different relation types.

\begin{table}[ht]
	\caption[]{Test set MRR for each relation in WN18RR. RotatE results are reported in \cite{zhang_quaternionknowledgegraph_2019}.}
	\label{tab:result_detailrelation}
	\centering  
	\begin{adjustbox}{max width=\linewidth}
		\begin{tabular}{@{\extracolsep{2pt}}lcc|c}
			
			Relation & RotatE & MEI & MEIM \\
			\midrule
			
			similar\_to & \textbf{1.000} & \textbf{1.000} & \textbf{1.000} \\ 
			hypernym & 0.148 & \underline{0.162} & \textbf{0.200} \\ 
			instance\_hypernym & 0.318 & \underline{0.405} & \textbf{0.411} \\ 
			member\_meronym & 0.232 & \underline{0.235} & \textbf{0.251} \\ 
			member\_of\_domain\_region & 0.200 & \underline{0.270} & \textbf{0.322} \\ 
			member\_of\_domain\_usage & \underline{0.318} & 0.295 & \textbf{0.352} \\ 
			synset\_domain\_topic\_of & 0.341 & \underline{0.359} & \textbf{0.418} \\ 
			also\_see & 0.585 & \underline{0.616} & \textbf{0.620} \\ 
			has\_part & 0.184 & \underline{0.203} & \textbf{0.211} \\ 
			verb\_group & 0.943 & \textbf{0.974} & \underline{0.969} \\ 
			derivationally\_related\_form & 0.947 & \textbf{0.959} & \underline{0.955} \\ 
			
			\midrule
			
			All relations & 0.476 & 0.481 & \textbf{0.499} \\
			
		\end{tabular}
	\end{adjustbox}
\end{table}

\subsection{Number of Partitions and Partition Size}
To examine the trade off between the core tensor and the embedding vectors, we keep the total number of parameters roughly the same and grid search different configurations on the FB15K-237 validation set. Table \ref{tab:param_tradeoff} shows that the MEIM model with 3 partitions each of size 100 achieves the best results. This is in agreement with previous theoretical prediction in the MEI paper \cite{tran_multipartitionembeddinginteraction_2020}. In practice, the best model should have the number of partitions $ K > 1 $ for ensemble effects, and the partition size C large enough for expressiveness but not too large for efficiency. As a rule of thumb, we can start with $ 3 \times 100 $ and grid search further.

\begin{table}[ht]
	\caption[]{The trade-off (K $ \times $ C) between the number of partitions (K) and partition size (C) on FB15K-237 validation set.}
	\label{tab:param_tradeoff}
	\centering  
	\begin{adjustbox}{max width=\linewidth}
		\begin{tabular}{@{\extracolsep{2pt}}cccc}
			
			Embedding size & \#params & MRR & H@10 \\
			\midrule
			
			$ 500 \times 1 $ \ \ \ \ & 7.39 M & 0.368 & 0.552 \\ 
			$ 250 \times 2 $ \ \ \ \ & 7.39 M & 0.367 & 0.549 \\ 
			$ 50 \times 10 $ & 7.44 M & 0.366 & 0.551 \\ 
			$ 20 \times 25 $ & 7.70 M & 0.369 & 0.556 \\ 
			\ \ $ 9 \times 50 $ & 7.78 M & 0.371 & 0.558 \\ 
			\ \ \ \ $ 3 \times 100 $ & 7.43 M & \textbf{0.375} & \textbf{0.562} \\ 
			\ \ \ \ $ 1 \times 170 $ & 7.43 M & 0.372 & 0.554 \\ 
			
		\end{tabular}
	\end{adjustbox}
\end{table}

\vfill\break

\subsection{Embedding sizes for comparable model sizes}
Table \ref{tab:comparable_size} shows the MEI embedding sizes used in this paper to make its model sizes comparable to that of MEIM. Please refer to the source code GitHub page for more information.

\begin{table}[ht]
	\centering  
	\begin{adjustbox}{max width=\linewidth}
		\begin{tabular}{@{\extracolsep{-4pt}}lccclccclccc}
			
			& \multicolumn{3}{c}{WN18RR} && \multicolumn{3}{c}{FB15K-237} && \multicolumn{3}{c}{YAGO3-10} \\
			\cmidrule{2-4} \cmidrule{6-8} \cmidrule{10-12}
			& K & C & \#params && K & C & \#params && K & C & \#params \\
			\cmidrule{2-4} \cmidrule{6-8} \cmidrule{10-12}
			
			MEIM & 3 & 100 & 15.3 M && 3 & 100 & 7.4 M && 5 & 100 & 66.6 M \\
			MEI & 3 & 115 & 15.7 M && 3 & 124 & 7.4 M && 5 & 106 & 66.5 M \\

		\end{tabular}
	\end{adjustbox}
	\caption[]{Embedding sizes for comparable MEI model sizes.}
	\label{tab:comparable_size}
\end{table}

\end{document}

%% file: math_commands.tex
\def\eqref#1{equation~\ref{#1}}

\def\1{\bm{1}}

\def\vh{{\bm{h}}}

\def\vr{{\bm{r}}}

\def\vt{{\bm{t}}}

\def\mH{{\bm{H}}}
\def\mI{{\bm{I}}}

\def\mM{{\bm{M}}}

\def\mR{{\bm{R}}}

\def\mT{{\bm{T}}}

\DeclareMathAlphabet{\mathsfit}{\encodingdefault}{\sfdefault}{m}{sl}
\SetMathAlphabet{\mathsfit}{bold}{\encodingdefault}{\sfdefault}{bx}{n}
\newcommand{\tens}[1]{\bm{\mathsfit{#1}}}

\def\tG{{\tens{G}}}

\def\tW{{\tens{W}}}

\def\gD{{\mathcal{D}}}
\def\gE{{\mathcal{E}}}

\def\gL{{\mathcal{L}}}

\def\gR{{\mathcal{R}}}
\def\gS{{\mathcal{S}}}

\def\sR{{\mathbb{R}}}

\newcommand{\diag}{{\text{diag}}} 

\theoremstyle{plain}

\theoremstyle{definition}

\theoremstyle{remark}




%% file: MEIM.bbl
\begin{thebibliography}{}

\bibitem[\protect\citeauthoryear{Bala{\v z}evi{\'c} \bgroup \em et al.\egroup
  }{2019}]{balazevic_tuckertensorfactorization_2019}
Ivana Bala{\v z}evi{\'c}, Carl Allen, and Timothy~M. Hospedales.
\newblock {{TuckER}}: {{Tensor Factorization}} for {{Knowledge Graph
  Completion}}.
\newblock In {\em Proceedings of the 2019 {{Conference}} on {{Empirical
  Methods}} in {{Natural Language Processing}}}, pages 5185--5194, 2019.

\bibitem[\protect\citeauthoryear{Bordes \bgroup \em et al.\egroup
  }{2013}]{bordes_translatingembeddingsmodeling_2013}
Antoine Bordes, Nicolas Usunier, Alberto {Garcia-Duran}, Jason Weston, and
  Oksana Yakhnenko.
\newblock Translating {{Embeddings}} for {{Modeling Multi-Relational Data}}.
\newblock In {\em Advances in {{Neural Information Processing Systems}}}, pages
  2787--2795, 2013.

\bibitem[\protect\citeauthoryear{Chami \bgroup \em et al.\egroup
  }{2020}]{chami_lowdimensionalhyperbolicknowledge_2020}
Ines Chami, Adva Wolf, Da-Cheng Juan, Frederic Sala, Sujith Ravi, and
  Christopher R{\'e}.
\newblock Low-{{Dimensional Hyperbolic Knowledge Graph Embeddings}}.
\newblock In {\em Proceedings of the 58th {{Annual Meeting}} of the
  {{Association}} for {{Computational Linguistics}}}, pages 6901--6914, 2020.

\bibitem[\protect\citeauthoryear{De~Lathauwer}{2008}]{delathauwer_decompositionshigherordertensor_2008a}
Lieven De~Lathauwer.
\newblock Decompositions of a {{Higher-Order Tensor}} in {{Block
  Terms}}\textemdash{{Part II}}: {{Definitions}} and {{Uniqueness}}.
\newblock {\em SIAM Journal on Matrix Analysis and Applications},
  30(3):1033--1066, 2008.

\bibitem[\protect\citeauthoryear{Dettmers \bgroup \em et al.\egroup
  }{2018}]{dettmers_convolutional2dknowledge_2018}
Tim Dettmers, Pasquale Minervini, Pontus Stenetorp, and Sebastian Riedel.
\newblock Convolutional {{2D Knowledge Graph Embeddings}}.
\newblock In {\em Proceedings of the 32nd {{AAAI Conference}} on {{Artificial
  Intelligence}}}, pages 1811--1818, 2018.

\bibitem[\protect\citeauthoryear{Ebisu and
  Ichise}{2019}]{ebisu_generalizedtranslationbasedembedding_2019}
Takuma Ebisu and Ryutaro Ichise.
\newblock Generalized {{Translation-based Embedding}} of {{Knowledge Graph}}.
\newblock {\em IEEE Transactions on Knowledge and Data Engineering},
  32(5):941--951, 2019.

\bibitem[\protect\citeauthoryear{Ha \bgroup \em et al.\egroup
  }{2016}]{ha_hypernetworks_2016}
David Ha, Andrew~M. Dai, and Quoc~V. Le.
\newblock {{HyperNetworks}}.
\newblock In {\em International {{Conference}} on {{Learning
  Representations}}}, page~18, 2016.

\bibitem[\protect\citeauthoryear{Ioffe and
  Szegedy}{2015}]{ioffe_batchnormalizationaccelerating_2015}
Sergey Ioffe and Christian Szegedy.
\newblock Batch {{Normalization}}: {{Accelerating Deep Network Training}} by
  {{Reducing Internal Covariate Shift}}.
\newblock In {\em International {{Conference}} on {{Machine Learning}}}, pages
  448--456, 2015.

\bibitem[\protect\citeauthoryear{Kazemi and
  Poole}{2018}]{kazemi_simpleembeddinglink_2018}
Seyed~Mehran Kazemi and David Poole.
\newblock {{SimplE Embedding}} for {{Link Prediction}} in {{Knowledge Graphs}}.
\newblock In {\em Advances in {{Neural Information Processing Systems}}}, pages
  4289--4300, 2018.

\bibitem[\protect\citeauthoryear{Kingma and
  Ba}{2015}]{kingma_adammethodstochastic_2015}
Diederik~P. Kingma and Jimmy Ba.
\newblock Adam: {{A Method}} for {{Stochastic Optimization}}.
\newblock In {\em International {{Conference}} on {{Learning
  Representations}}}, page~15, 2015.

\bibitem[\protect\citeauthoryear{Kolda and
  Bader}{2009}]{kolda_tensordecompositionsapplications_2009}
Tamara~G. Kolda and Brett~W. Bader.
\newblock Tensor {{Decompositions}} and {{Applications}}.
\newblock {\em SIAM Review}, 51(3):455--500, 2009.

\bibitem[\protect\citeauthoryear{Liu \bgroup \em et al.\egroup
  }{2021}]{liu_ragatrelationaware_2021}
Xiyang Liu, Huobin Tan, Qinghong Chen, and Guangyan Lin.
\newblock {{RAGAT}}: {{Relation Aware Graph Attention Network}} for {{Knowledge
  Graph Completion}}.
\newblock {\em IEEE Access}, 9:20840--20849, 2021.

\bibitem[\protect\citeauthoryear{Mahdisoltani \bgroup \em et al.\egroup
  }{2015}]{mahdisoltani_yago3knowledgebase_2015}
Farzaneh Mahdisoltani, Joanna Biega, and Fabian~M. Suchanek.
\newblock {{YAGO3}}: {{A Knowledge Base}} from {{Multilingual Wikipedias}}.
\newblock In {\em Proceedings of the {{Conference}} on {{Innovative Data
  Systems Research}}}, 2015.

\bibitem[\protect\citeauthoryear{Mason \bgroup \em et al.\egroup
  }{1999}]{mason_boostingalgorithmsgradient_1999}
Llew Mason, Jonathan Baxter, Peter Bartlett, and Marcus Frean.
\newblock Boosting {{Algorithms}} as {{Gradient Descent}}.
\newblock In {\em Advances in {{Neural Information Processing Systems}}}, pages
  512--518, 1999.

\bibitem[\protect\citeauthoryear{Nickel \bgroup \em et al.\egroup
  }{2011}]{nickel_threewaymodelcollective_2011}
Maximilian Nickel, Volker Tresp, and Hans-Peter Kriegel.
\newblock A {{Three-Way Model}} for {{Collective Learning}} on
  {{Multi-Relational Data}}.
\newblock In {\em International {{Conference}} on {{Machine Learning}}}, pages
  809--816, 2011.

\bibitem[\protect\citeauthoryear{Rossi \bgroup \em et al.\egroup
  }{2021}]{rossi_knowledgegraphembedding_2021}
Andrea Rossi, Denilson Barbosa, Donatella Firmani, Antonio Matinata, and Paolo
  Merialdo.
\newblock Knowledge graph embedding for link prediction: {{A}} comparative
  analysis.
\newblock {\em ACM Transactions on Knowledge Discovery from Data (TKDD)},
  15(2):1--49, 2021.

\bibitem[\protect\citeauthoryear{Ruffinelli \bgroup \em et al.\egroup
  }{2020}]{ruffinelli_youcanteach_2020}
Daniel Ruffinelli, Samuel Broscheit, and Rainer Gemulla.
\newblock You {{CAN Teach}} an {{Old Dog New Tricks}}! {{On Training Knowledge
  Graph Embeddings}}.
\newblock In {\em International {{Conference}} on {{Learning
  Representations}}}, page~20, 2020.

\bibitem[\protect\citeauthoryear{Sok{\'o}{\l} and
  Park}{2020}]{sokol_informationgeometryorthogonal_2020}
Piotr~Aleksander Sok{\'o}{\l} and Il~Memming Park.
\newblock Information {{Geometry}} of {{Orthogonal Initializations}} and
  {{Training}}.
\newblock In {\em International {{Conference}} on {{Learning
  Representations}}}, page~17, 2020.

\bibitem[\protect\citeauthoryear{Srivastava \bgroup \em et al.\egroup
  }{2014}]{srivastava_dropoutsimpleway_2014}
Nitish Srivastava, Geoffrey Hinton, Alex Krizhevsky, Ilya Sutskever, and Ruslan
  Salakhutdinov.
\newblock Dropout: {{A Simple Way}} to {{Prevent Neural Networks}} from
  {{Overfitting}}.
\newblock {\em The Journal of Machine Learning Research}, 15(1):1929--1958,
  2014.

\bibitem[\protect\citeauthoryear{Sun \bgroup \em et al.\egroup
  }{2019}]{sun_rotateknowledgegraph_2019}
Zhiqing Sun, Zhi-Hong Deng, Jian-Yun Nie, and Jian Tang.
\newblock {{RotatE}}: {{Knowledge Graph Embedding}} by {{Relational Rotation}}
  in {{Complex Space}}.
\newblock In {\em International {{Conference}} on {{Learning
  Representations}}}, page~18, 2019.

\bibitem[\protect\citeauthoryear{Sun \bgroup \em et al.\egroup
  }{2020}]{sun_reevaluationknowledgegraph_2020}
Zhiqing Sun, Shikhar Vashishth, Soumya Sanyal, Partha Talukdar, and Yiming
  Yang.
\newblock A {{Re-evaluation}} of {{Knowledge Graph Completion Methods}}.
\newblock In {\em Proceedings of the 58th {{Annual Meeting}} of the
  {{Association}} for {{Computational Linguistics}}}, pages 5516--5522, 2020.

\bibitem[\protect\citeauthoryear{Tang \bgroup \em et al.\egroup
  }{2020}]{tang_orthogonalrelationtransforms_2020}
Yun Tang, Jing Huang, Guangtao Wang, Xiaodong He, and Bowen Zhou.
\newblock Orthogonal {{Relation Transforms}} with {{Graph Context Modeling}}
  for {{Knowledge Graph Embedding}}.
\newblock In {\em Proceedings of the {{Annual Meeting}} of the {{Association}}
  for {{Computational Linguistics}}}, pages 2713--2722, 2020.

\bibitem[\protect\citeauthoryear{Toutanova and
  Chen}{2015}]{toutanova_observedlatentfeatures_2015}
Kristina Toutanova and Danqi Chen.
\newblock Observed versus latent features for knowledge base and text
  inference.
\newblock In {\em Proceedings of the 3rd {{Workshop}} on {{Continuous Vector
  Space Models}} and Their {{Compositionality}}}, pages 57--66, 2015.

\bibitem[\protect\citeauthoryear{Tran and
  Takasu}{2019a}]{tran_analyzingknowledgegraph_2019}
Hung-Nghiep Tran and Atsuhiro Takasu.
\newblock Analyzing {{Knowledge Graph Embedding Methods}} from a
  {{Multi-Embedding Interaction Perspective}}.
\newblock In {\em Proceedings of the {{Data Science}} for {{Industry}} 4.0
  {{Workshop}} at {{EDBT}}/{{ICDT}}}, page~7, 2019.

\bibitem[\protect\citeauthoryear{Tran and
  Takasu}{2019b}]{tran_exploringscholarlydata_2019}
Hung-Nghiep Tran and Atsuhiro Takasu.
\newblock Exploring {{Scholarly Data}} by {{Semantic Query}} on {{Knowledge
  Graph Embedding Space}}.
\newblock In {\em Proceedings of the 23rd {{International Conference}} on
  {{Theory}} and {{Practice}} of {{Digital Libraries}}}, pages 154--162, 2019.

\bibitem[\protect\citeauthoryear{Tran and
  Takasu}{2020}]{tran_multipartitionembeddinginteraction_2020}
Hung-Nghiep Tran and Atsuhiro Takasu.
\newblock Multi-{{Partition Embedding Interaction}} with {{Block Term Format}}
  for {{Knowledge Graph Completion}}.
\newblock In {\em Proceedings of the {{European Conference}} on {{Artificial
  Intelligence}}}, pages 833--840, 2020.

\bibitem[\protect\citeauthoryear{Tran}{2020}]{tran_multirelationalembeddingknowledge_2020}
Hung-Nghiep Tran.
\newblock {\em Multi-{{Relational Embedding}} for {{Knowledge Graph
  Representation}} and {{Analysis}}}.
\newblock PhD thesis, The Graduate University for Advanced Studies, SOKENDAI,
  {Japan}, 2020.

\bibitem[\protect\citeauthoryear{Trouillon \bgroup \em et al.\egroup
  }{2016}]{trouillon_complexembeddingssimple_2016}
Theo Trouillon, Johannes Welbl, Sebastian Riedel, {Eric Gaussier}, and
  {Guillaume Bouchard}.
\newblock Complex {{Embeddings}} for {{Simple Link Prediction}}.
\newblock In {\em International {{Conference}} on {{Machine Learning}}}, pages
  2071--2080, 2016.

\bibitem[\protect\citeauthoryear{Tucker}{1966}]{tucker_mathematicalnotesthreemode_1966}
Ledyard~R Tucker.
\newblock Some {{Mathematical Notes}} on {{Three-Mode Factor Analysis}}.
\newblock {\em Psychometrika}, 31(3):279--311, 1966.

\bibitem[\protect\citeauthoryear{Vashishth \bgroup \em et al.\egroup
  }{2020a}]{vashishth_interacteimprovingconvolutionbased_2020}
Shikhar Vashishth, Soumya Sanyal, Vikram Nitin, Nilesh Agrawal, and Partha
  Talukdar.
\newblock {{InteractE}}: {{Improving Convolution-based Knowledge Graph
  Embeddings}} by {{Increasing Feature Interactions}}.
\newblock In {\em Proceedings of the 34th {{AAAI Conference}} on {{Artificial
  Intelligence}}}, pages 3009--3016, 2020.

\bibitem[\protect\citeauthoryear{Vashishth \bgroup \em et al.\egroup
  }{2020b}]{vashishth_compositionbasedmultirelationalgraph_2020}
Shikhar Vashishth, Soumya Sanyal, Vikram Nitin, and Partha Talukdar.
\newblock Composition-based {{Multi-Relational Graph Convolutional Networks}}.
\newblock In {\em International {{Conference}} on {{Learning
  Representations}}}, page~16, 2020.

\bibitem[\protect\citeauthoryear{Yang \bgroup \em et al.\egroup
  }{2015}]{yang_embeddingentitiesrelations_2015}
Bishan Yang, Wen-tau Yih, Xiaodong He, Jianfeng Gao, and Li~Deng.
\newblock Embedding {{Entities}} and {{Relations}} for {{Learning}} and
  {{Inference}} in {{Knowledge Bases}}.
\newblock In {\em International {{Conference}} on {{Learning
  Representations}}}, page~12, 2015.

\bibitem[\protect\citeauthoryear{Zhang \bgroup \em et al.\egroup
  }{2019}]{zhang_quaternionknowledgegraph_2019}
Shuai Zhang, Yi~Tay, Lina Yao, and Qi~Liu.
\newblock Quaternion {{Knowledge Graph Embedding}}.
\newblock In {\em Advances in {{Neural Information Processing Systems}}}, pages
  2735--2745, 2019.

\end{thebibliography}
